\documentclass[10pt,twocolumn,letterpaper]{article}

\usepackage{cvpr}              
\makeatletter
\@namedef{ver@everyshi.sty}{}
\makeatother
\usepackage{tikz}

\usepackage[accsupp]{axessibility}  
\usepackage{graphicx}
\usepackage{amsmath}
\usepackage{amssymb}
\usepackage{booktabs}
\usepackage{gensymb}
\usepackage{bbm}
\usepackage{multirow}
\usepackage{array}
\usepackage{float}

\usepackage{changes}

\usepackage[pagebackref,breaklinks,colorlinks]{hyperref}

\usepackage[capitalize]{cleveref}
\crefname{section}{Sec.}{Secs.}
\Crefname{section}{Section}{Sections}
\Crefname{table}{Table}{Tables}
\crefname{table}{Tab.}{Tabs.}

\begin{document}

\title{BKinD-3D: Self-Supervised 3D Keypoint Discovery from Multi-View Videos}

\author{Jennifer J. Sun\thanks{Equal contribution}\\
Caltech 
\and 
Lili Karashchuk\footnotemark[1] \\
U Washington
\and
Amil Dravid\footnotemark[1] \\
Northwestern 
\and 
Serim Ryou\\ 
SAIT\thanks{Work done outside of SAIT} 
\and
Sonia Fereidooni \\
U Washington 
\and
John C. Tuthill \\
U Washington 
\and
Aggelos Katsaggelos\\
Northwestern  
\and
Bingni W. Brunton \\
U Washington 
\and
Georgia Gkioxari \\
Caltech 
\and
Ann Kennedy\\
Northwestern  
\and 
Yisong Yue\\
Caltech 
\and 
Pietro Perona\\
Caltech 
\and
{\small Code \& Project Website: \url{https://sites.google.com/view/b-kind/3d}}
}

\maketitle

\begin{abstract}
\vspace{-0.1in}
Quantifying motion in 3D is important for studying the behavior of humans and other animals, but manual pose annotations are expensive and time-consuming to obtain.
Self-supervised keypoint discovery is a promising strategy for estimating 3D poses without annotations.
However, current keypoint discovery approaches commonly process single 2D views and do not operate in the 3D space.
We propose a new method to perform self-supervised keypoint discovery in 3D from multi-view videos of behaving agents, without any keypoint or bounding box supervision in 2D or 3D. 
Our method, BKinD-3D, uses an encoder-decoder architecture with a 3D volumetric heatmap, trained to reconstruct spatiotemporal differences across multiple views, in addition to joint length constraints on a learned 3D skeleton of the subject.
In this way, we discover keypoints without requiring manual supervision in videos of humans and rats, demonstrating the potential of 3D keypoint discovery for studying behavior.
\vspace{-0.1in}
\end{abstract}

\section{Introduction}\label{sec:intro}

All animals behave in 3D, and analyzing 3D posture and movement is crucial for a variety of applications, including the study of biomechanics, motor control, and behavior~\cite{marshall2022leaving}.
However, annotations for supervised training of 3D pose estimators are expensive and time-consuming to obtain, especially for studying diverse animal species and varying experimental contexts.
Self-supervised keypoint discovery has demonstrated tremendous potential in discovering 2D keypoints from video~\cite{JakabNeurips18,jakab20self-supervised,sun2022self}, without the need for manual annotations.
These models have not been well-explored in 3D, which is more challenging compared to 2D due to depth ambiguities, a larger search space, and the need to incorporate geometric constraints.
Our goal is to enable 3D keypoint discovery of humans and animals from synchronized multi-view videos, without 2D or 3D supervision.

\begin{figure}
    \centering
    \vspace{-0.1in}
    \includegraphics[width=\linewidth]{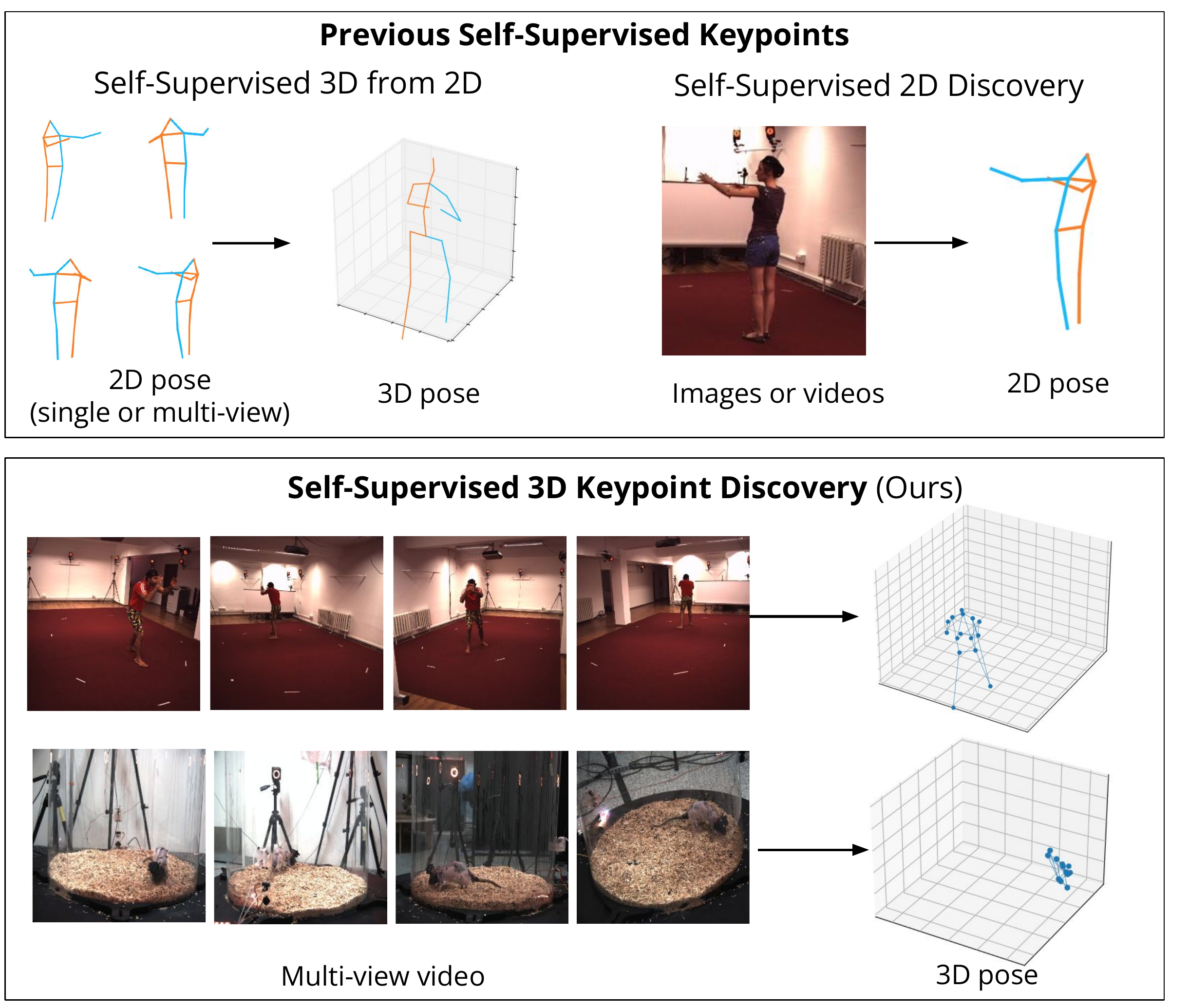}
    \caption{\textbf{Self-supervised 3D keypoint discovery}. Previous work studying self-supervised keypoints either requires 2D supervision for 3D pose estimation or focuses on 2D keypoint discovery. Currently, self-supervised 3D keypoint discovery is not well-explored. We propose methods for discovering 3D keypoints directly from multi-view videos of different organisms, such as human and rats, without 2D or 3D supervision. The 3D keypoint discovery examples demonstrate the results from our method. }
      \vspace{-0.1in}
    \label{fig:intro}
\end{figure}

\textbf{Self-Supervised 3D Keypoint Discovery.}  
Previous works for self-supervised 3D keypoints typically start from a pre-trained 2D pose estimator~\cite{usman2022metapose,kocabas2019self}, and thus do not perform \textit{keypoint discovery} (Figure~\ref{fig:intro}). 
These models are suitable for studying human poses because 2D human pose estimators are widely available and the pose and body structure of humans is well-defined. 
However, for many scientific applications~\cite{pereira2020quantifying,marshall2022leaving,sun2022self}, it is important to track diverse organisms in different experimental contexts. These situations require time-consuming 2D or 3D annotations for training pose estimation models.
The goal of our work is to enable 3D keypoint discovery from multi-view videos directly, without any 2D or 3D supervision, in order to accelerate the analysis of 3D poses from diverse animals in novel settings.
To the best of our knowledge, self-supervised 3D keypoint discovery have not been well-explored for real-world multi-view videos.

\textbf{Behavioral Videos.} We study 3D keypoint discovery in the setting of behavioral videos with stationary cameras and backgrounds.
We chose this for several reasons.
First, this setting is common in many real-world behavior analysis datasets~\cite{segalin2020mouse,eyjolfsdottir2014detecting,burgos2012social,marstaller2019deepbees,pereira2020quantifying,jhuang2010automated,sun2021multi}, where there has been an emerging trend to expand the study of behavior from 2D to 3D~\cite{marshall2022leaving}. 
Thus, 3D keypoint discovery would directly benefit many scientific studies in this space using approaches such as biomechanics, motor control, and behavior~\cite{marshall2022leaving}.
Second, studying behavioral videos in 3D enables us to leverage recent work in 2D keypoint discovery for behavioral videos~\cite{sun2022self}. 
Finally, this setting enables us to tackle the 3D keypoint discovery challenge in a modular way. 
For example, in behavior analysis experiments, many tools are already available for camera calibration~\cite{karashchuk2021anipose}, and we can assume that camera parameters are known. 

\textbf{Our Approach.} The key to our approach, which we call \textbf{B}ehavioral \textbf{K}eypo\textbf{in}t
\textbf{D}iscovery in \textbf{3D} (BKinD-3D), is to encode self-supervised learning signals from videos across multiple views into a single 3D geometric bottleneck.
We leverage the spatiotemporal difference reconstruction loss from~\cite{sun2022self} and use multi-view reconstruction to train an encoder-decoder architecture. 
Our method does not use any bounding boxes or keypoint annotations as supervision.
Critically, we impose links between our discovered keypoints to discover connectivity across points.
In other words, keypoints on the same parts of the body are connected, so that we are able to enforce joint length constraints in 3D.
To show that our model is applicable across multiple settings, we demonstrate our approach on multi-view videos from different organisms. To summarize:

\begin{table}
  \begin{center}
\scalebox{0.8}{
    \begin{tabular}{lcccccc}
        \toprule[0.2em]
        Method & 3D sup. & 2D sup. & camera params & data type \\
        \toprule[0.2em]
        Isakov et al.~\cite{iskakov2019learnable} & \multirow{2}{*}{\checkmark} & \multirow{2}{*}{\checkmark} & intrinsics & \multirow{2}{*}{real} \\
       DANNCE~\cite{dunn2021geometric} & & & extrinsics & \\ 
        \hline
        Rhodin et al.~\cite{rhodin2018learning} & \checkmark &  optional  &  intrinsics & real\\    
         \hline
        Anipose~\cite{karashchuk2021anipose} & \multirow{2}{*}{$\times$} & \multirow{2}{*}{\checkmark} &   intrinsics & \multirow{2}{*}{real}\\
        DeepFly3D~\cite{gunel_deepfly3d_2019} & & & extrinsics & \\  
        \hline
        EpipolarPose~\cite{kocabas2019self} &  \multirow{2}{*}{$\times$}  & \multirow{2}{*}{\checkmark} & \multirow{2}{*}{optional} & \multirow{2}{*}{real}\\       
        CanonPose~\cite{wandt2021canonpose} & & & & \\
        \hline
        MetaPose~\cite{usman2022metapose} &  $\times$  & \checkmark & $\times$ & real\\       
        \hline
        \multirow{2}{*}{Keypoint3D~\cite{chen2021unsupervised}} &  \multirow{2}{*}{$\times$}  &  \multirow{2}{*}{$\times$} &  intrinsics & \multirow{2}{*}{simulation}\\
        & & & extrinsics & \\       
        \hline
        \multirow{2}{*}{Ours (3D discovery)} &  \multirow{2}{*}{$\times$}  &  \multirow{2}{*}{$\times$}  &  intrinsics & \multirow{2}{*}{real} \\    
        & & & extrinsics & \\        
        \bottomrule[0.1em]
    \end{tabular}}
  \caption{\textbf{Comparison of our work with representative related work for 3D pose using multi-view training}. Previous works require either 3D or 2D supervision, or simulated environments to train jointly with reinforcement learning. Our method addresses a gap in discovering 3D keypoints from real videos without 2D or 3D supervision.}
  \vspace{-0.2in}
  \label{tab:related_work}
  \end{center}
\end{table}

\begin{itemize}
    \item We introduce self-supervised 3D keypoint discovery, which discovers 3D pose from real-world multi-view behavioral videos of different organisms, without any 2D or 3D supervision. 
    \item We propose a novel method (BKinD-3D) for end-to-end 3D discovery from video using multi-view spatiotemporal difference reconstruction and 3D joint length constraints. 
    \item We demonstrate quantitatively that our work significantly closes the gap between supervised 3D methods and 3D keypoint discovery across different organisms (humans and rats). 
\end{itemize}


\section{Related Work}

\textbf{3D Pose Estimation.} There has been a large body of work studying 3D human pose estimation from images or videos, as reviewed in~\cite{sarafianos20163d,wang2021deep}, with recent works also focusing on 3D animal poses~\cite{dunn2021geometric,marshall2022leaving,gosztolai2021liftpose3d,karashchuk2021anipose,gunel_deepfly3d_2019}. 
Most of these methods are fully supervised from visual data~\cite{iskakov2019learnable,sun2018integral,chen2020cross}, with some models perform lifting starting from 2D poses~\cite{martinez2017simple,chen20173d,pavllo20193d,rayat2018exploiting}.
We focus our discussion on multi-view 3D pose estimation methods, but all of these models require either 3D or 2D supervision during training. 
This 2D supervision is typically in the form of pre-trained 2D detectors~\cite{kocabas2019self}, or ground truth 2D poses~\cite{usman2022metapose}.
In comparison, our method uses multi-view videos to discover 3D keypoints without 2D or 3D supervision.

Methods more closely related to our work are those that also leverage multi-view structure to estimate 3D pose (Table~\ref{tab:related_work}). 
\cite{iskakov2019learnable} proposed a supervised method that uses learnable triangulation to aggregate 2D information across views to 3D. 
Here we study similar approaches for representing 3D information, but using self-supervision instead of supervised 3D annotations.
Other methods in this space propose training methods such as enforcing consistency of predicted poses across views~\cite{rhodin2018learning}, regression to 3D pose estimated from epipolar geometry of multi-view 2D~\cite{kocabas2019self}, constraining 3D poses to project to realistic 2D pose~\cite{chen2019unsupervised}, or estimates camera parameters using detected and ground truth 2D poses~\cite{usman2022metapose}.
While we also leverage multi-view information, our goal is different from the work above, in that our approach aims to discover 3D poses without 2D or 3D supervision, given camera parameters.

\begin{figure*}
    \centering
    \vspace{-0.1in}
    \includegraphics[width=\linewidth]{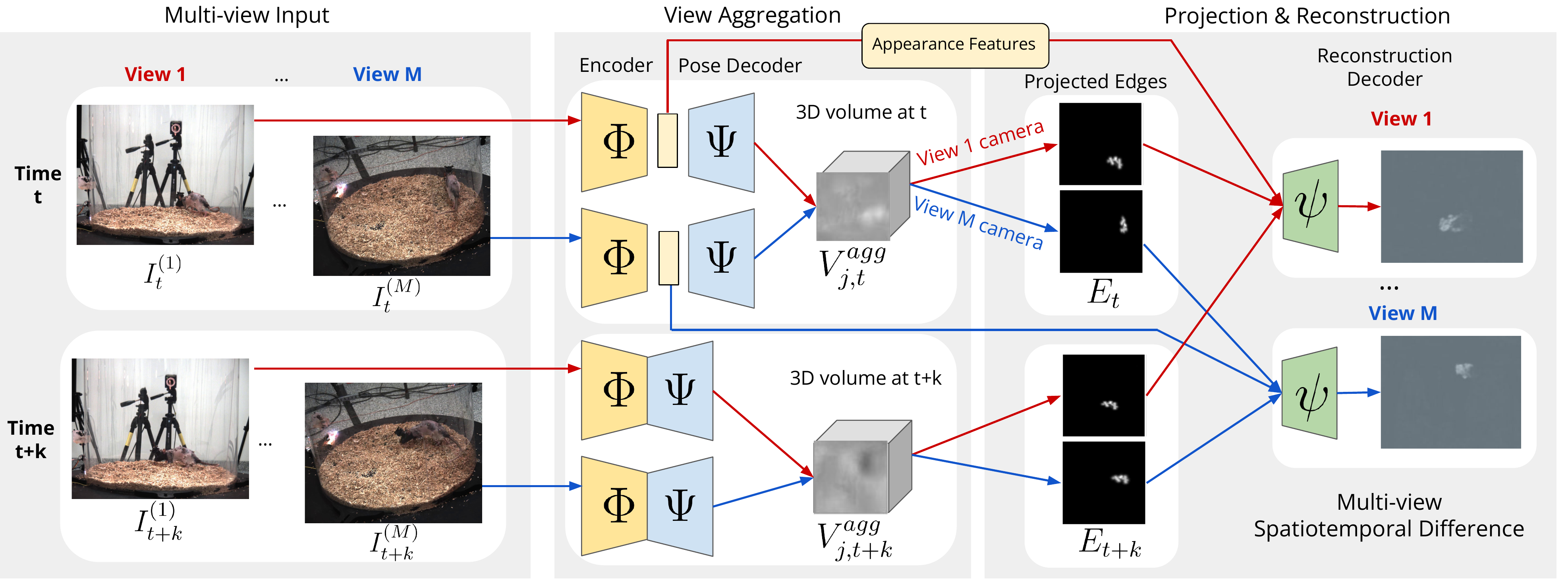}
    \caption{\textbf{BKinD-3D: 3D keypoint discovery using 3D volume bottleneck}. We start from input multi-view videos with known camera parameters, then unproject feature maps from geometric encoders into 3D volumes for timestamps $t$ and $t+k$. We next aggregate 3D points from volumes into a single edge map at each timestamp, and use edges as input to the decoder alongside appearance features at time $t$. The model is trained using multi-view spatiotemporal difference reconstruction. Best viewed in color.
    \vspace{-0.1in}
    }
    \label{fig:method}
\end{figure*}

\textbf{Self-supervised Keypoint Discovery.} 
2D keypoint discovery has been studied from images~\cite{JakabNeurips18,ZhangKptDisc18,he2022autolink} and videos~\cite{jakab20self-supervised,sun2022self}. 
Our approach focuses on behavioral videos, similar to~\cite{sun2022self}, but we aim to use multi-view information to discover 3D keypoints, instead of 2D.
Many approaches use an encoder-decoder setup to disentangle appearance and geometry information~\cite{ZhangKptDisc18,JakabNeurips18,Lorenz19,sun2022self}.
Our setup also consists of encoders and decoders, but our encoder maps information across views to aggregate 2D information into a 3D geometry bottleneck.
The discovery model most similar to our approach is Keypoint3D~\cite{chen2021unsupervised}, which discovers 3D keypoints for control from virtual agents, using a combination of image reconstruction and reinforcement learning.
However, this setup is designed for simulated data and does not translate well to real videos, since updating the keypoints through a reinforcement learning policy requires videos generated through the simulated environment. 
Keypoint discovery models typically represent discovered parts as 2D Gaussian heatmaps~\cite{JakabNeurips18,sun2022self} or 2D edges~\cite{he2022autolink}.
While we also use an edge-based representation, our edges are in 3D, which enables our training objective to enforce joint length consistency.

\textbf{Behavioral Video Analysis.}
Pose estimation is a common intermediate step in automated behavior quantification; behavioral videos are commonly captured with stationary camera and background, with moving agents.  
To date, supervised 2D pose estimators are most often used for analyzing behavior videos~\cite{kabra2013jaaba,hong2015automated,eyjolfsdottir2016learning,Mathisetal2018,egnor2016computational,segalin2020mouse}.
However, 2D pose estimation is inadequate for many applications: it cannot reliably capture the angle of joints for kinematics, fails to generalize across views, is sensitive to occlusion, and cannot incorporate body plan constraints as skeleton length or range of motion of joints.
Thus, there has recently been an accelerating trend to study behavior in 3D~\cite{karashchuk2021anipose,marshall2022leaving,dunn2021geometric,gosztolai2021liftpose3d}.
These models typically require more expensive 3D training annotations compared to 2D poses.
While 2D self-supervision has been studied for behavioral videos~\cite{sun2022self}, 3D keypoint discovery in real-world behavioral videos have not been well-explored.

\section{Method}\label{sec:method}

Our goal is to discover 3D keypoints from multi-view behavioral videos without 2D or 3D supervision (Figure~\ref{fig:method}).
Our approach is inspired by BKinD~\cite{sun2022self}, which uses spatiotemporal difference reconstruction to discover 2D keypoints in behavioral videos. 
In these videos, the camera and background is stationary, and spatiotemporal difference provides a strong signal for encoding agent movement.

We develop several approaches for 3D keypoint discovery, but focus on our volumetric model (Figure~\ref{fig:method}) in this section, as this model generally performed the best in our evaluations. 
More details on other approaches are in Section~\ref{sec:model_comparisons} and supplemental materials.

In our volumetric model (BKinD-3D, Figure~\ref{fig:method}) we use multi-view spatiotemporal reconstruction to train an encoder-decoder architecture with 2D information aggregated to a 3D volumetric heatmap. Projections from the 3D heatmap in the form of agent skeletons are then used to reconstruct movement, represented by spatiotemporal difference, in each view.

\subsection{3D Keypoint Discovery}

Given behavioral videos captured from $M$ synchronized camera views, with known camera projection matrix $P^{(i)}$ for each camera $i \in
\{1...M \}$, we aim to discover a set of $J$ 3D keypoints $U_t \in \mathbb{R}^{J \times 3}$ on a single behaving agent, at each timestamp $t$. We assume access to camera projection matrices so that our model discovers 3D keypoints in the global coordinate frame.

During training, our model uses two timestamps in the video $t$ and $t+k$ to compute the spatiotemporal difference in each view as the reconstruction target. In other words, for each camera view $i$, our training starts with a frame $I_t^{(i)}$ and a future frame $I_{t+k}^{(i)}$. During inference, only a single timestamp is required:
once the model is trained, the model only needs $I_t^{(i)}$ for each camera view $i$.

In our model setup, the appearance encoder $\Phi$, geometry decoder $\Psi$, and reconstruction decoder $\psi$ are shared across views and timestamps (in previous work~\cite{sun2022self}, these networks are shared across timestamps, but only a single view is addressed). The appearance encoder $\Phi$ is used to generate appearance features, which are decoded into 2D heatmaps by the geometry decoder $\Psi$. These 2D heatmaps are then aggregated across views to form a 3D volumetric bottleneck (Section~\ref{sec:view_agg}), which is processed by a volume-to-volume network $\rho$. We compute the 3D keypoints using spatial softmax on the 3D volume. Then, we project these keypoints to 2D, compute edges between points, and output these edges into the reconstruction decoder $\psi$ (Section~\ref{sec:projection_recon}) for training. The reconstruction decoder $\psi$ is only used during training, and not required for inference.
\vspace{-0.1in}

\subsubsection{Feature Encoding}
To start, we first compute appearance features from frame pairs $I_t^{(i)}$ and $I_{t+k}^{(i)}$ using the appearance encoder $\Phi$: $\Phi(I_t^{(i)})$ and $\Phi(I_{t+k}^{(i)})$. These appearance features are then fed into the geometry decoder $\Psi$ to generate 2D heatmaps $\Psi(\Phi(I_t^{(i)})) = H_t^{(i)}$ and $H_{t+k}^{(i)}$.
Each 2D heatmap has $C$ channels, where $H_{t,c}^{(i)}$ represents channel $c$ of $H_t^{(i)}$.

\subsubsection{View Aggregation using Volumetric Model}\label{sec:view_agg}

To aggregate information across views, we unproject our 2D heatmaps to a 3D volumetric bottleneck.
We perform view aggregation separately across timestamps $t$ and $t+k$.

We aggregate 2D heatmaps into a 3D volume similar to~\cite{iskakov2019learnable}, which used previously for supervised 3D human pose estimation.
One important difference is that in the supervised setting, an $L \times L \times L$ sized volume is drawn around the human pelvis, with $L$ being around twice the size of a person.
As we perform keypoint discovery, we do not have information on the location or size of the agent. 
Instead, we initialize our volume with $L$ representing the maximum size of the space/room for the behaving agent.

This process aggregates 2D heatmaps $H_{t,c}^{(i)}$ for cameras $i \in \{1...M\}$ and channels $c \in \{1...C\}$ to 3D keypoints $U_t$, for timestamp $t$.
Our volume is first discretized into voxels $V_{coords} \in \mathbb{R}^{B \times B \times B \times 3}$, where $B$ represents the number of distinct coordinates in each dimension. Each voxel corresponds to a global 3D coordinate.
These 3D coordinates are projected to a 2D plane using the projection matrices in each camera view $i$: $V_{proj}^{(i)} = P^{(i)} V_{coords}$.
A volume $V_{c}^{(i)}$ is then created and filled for each camera view $i$ and each channel $c$ using bilinear sampling~\cite{jaderberg2015spatial} from the corresponding 2D heatmap: $V_{c}^{(i)} = H_{t,c}^{(i)}\{V_{proj}^{(i)}\}$, where $\{\cdot\}$ denotes bilinear sampling.

We then aggregate these $V_{c}^{(i)}$ across views for each channel $c$ using a softmax approach~\cite{iskakov2019learnable}:
\vspace{-0.05in}
$$V_c^{agg} = \sum_{i} \frac{\exp(V_{c}^{(i)})}{\sum_j \exp(V_{c}^{(j)})} \odot V_{c}^{(i)}.$$
$V^{agg}$ is then mapped to 3D heatmaps corresponding to each joint using a volumetric convolutional network~\cite{moon2018v2v} $\rho$: $V^{agg*} = \rho(V^{agg})$. We compute the 3D spatial softmax over the volume, for each channel $j$ of $V_j^{agg*}$, $j \in \{1...J\}$, to obtain the 3D keypoint locations $U_{t}$ for timestamp t, as in \cite{iskakov2019learnable}. 
In many supervised works, the keypoint locations $U_{t}$ are optimized to match to ground truth 3D poses; however, we aim to discover 3D keypoints, and train our network by using $U_{t}$ to decode spatiotemporal difference across views.

\vspace{-0.05in}
\subsubsection{Projection and Reconstruction}\label{sec:projection_recon}

In this step, we project the discovered 3D keypoints to a 2D representation in each view using camera parameters. For training, 2D representations in timestamps $t$ and $t+k$ are used as input to the reconstruction decoder $\psi$.
We train the 3D keypoints $U_t$ at each timestamp $t$ using multi-view spatiotemporal difference reconstruction.
The target spatiotemporal difference is computed using the 2D image pair $I_t^{(i)}$ and $I_{t+k}^{(i)}$ at each view $i$. 

First, we project the 3D keypoints using camera projection matrices into 2D keypoints $u_t^{(i)} = P^{(i)} U_t$.
We create an edge representation for each view for each timestamp, which enables us to discover connections between points and enforce 3D joint length constraints.
For each keypoint pair $u_{t,m}^{(i)}$ and $u_{t,n}^{(i)}$, we draw a differentiable edge map as a Gaussian along the line connecting them, similar to~\cite{he2022autolink}:
\vspace{-0.05in}
$$E_{t, (m,n)}^{(i)}(\mathbf{p}) = \exp(d_{m,n}^{(i)}(\mathbf{p})^2/\sigma^2),$$
where $\sigma$ controls the line thickness and $d_{m,n}(\mathbf{p})^{(i)}$ is the distance between pixel $\mathbf{p}$ and the line connecting $u_{t,m}^{(i)}$ and $u_{t,n}^{(i)}$. We then aggregate the edge heatmaps at each timestamp using a set of learned weights $w_{m,n}$ for each edge, where $w_{m,n}$ is shared across all timestamps and all views. An edge is active and connects two points if $w_{m,n} > 0$, otherwise the points are not connected.
Finally, we aggregate all edge heatmaps using the max across all edge pairs~\cite{he2022autolink}: $$E_t^{(i)}(\mathbf{p}) = \max_{m,n} w_{m,n} E_{t, (m,n)^{(i)}}(\mathbf{p}).$$

In our framework, for each view $i$, the decoder $\psi$ uses the edge maps $E_t^{(i)}$ and $E_{t+k}^{(i)}$ as well as the appearance feature $\Phi(I_t^{(i)})$ for reconstructing the spatiotemporal difference across each view. 
The ground truth spatiotemporal difference is computed from the original images $S(I_t^{(i)}, I_{t+k}^{(i)})$.
The reconstruction from the model is $\hat{S} = \psi(E_t^{(i)}, E_{t+k}^{(i)}, \Phi(I_t^{(i)}))$, through the 3D volumetric bottleneck in order to discover informative 3D keypoints for reconstructing agent movement.

\subsection{Learning Formulation}

The entire training pipeline (Figure~\ref{fig:method}) is differentiable, and we train the model end-to-end. We note that our model is only given multi-view video and corresponding camera parameters, without keypoint or bounding box supervision. 

\vspace{-0.05in}
\subsubsection{Multi-View Reconstruction Loss}

Our multi-view spatiotemporal difference reconstruction is based on the single-view spatiotemporal difference studied for 2D keypoint discovery~\cite{sun2022self}. We compute the Structural Similarity Index Measure (SSIM)~\cite{Wang04imagequality} as a reconstruction target in each view. 
SSIM has been used to measure perceived differences between images based on luminance, contrast, and structure features. 
Here, we use SSIM as a reconstruction target and we compute a similarity map using local SSIM on corresponding patches between $I_t^{(i)}$ and $I_{t+k}^{(i)}$. This similarity map is negated to obtain the dissimilarity map used as the target: $S(I_t^{(i)}, I_{t+k}^{(i)})$.

We use perceptual loss~\cite{Johnson2016Perceptual} in each view between the target $S$ and the reconstruction $\hat{S}$. This loss computes the L2 distance between features of the target and reconstruction computed from the VGG network $\phi$~\cite{VGG14}:
\begin{align}
    \mathcal{L}_{recon}^{(i)} = \left\Vert \phi(S(I_t^{(i)},I_{t+T}^{(i)})) - \phi(\hat{S}(I_t^{(i)},I_{t+T}^{(i)})) \right\Vert_2.
\end{align}
The error is computed by comparing features from intermediate convolutional blocks of the network. Our final perceptual loss is summed over each view $\mathcal{L}_{recon} = \sum_i \mathcal{L}_{recon}^{(i)}$.

\vspace{-0.05in}
\subsubsection{Learned Length Constraint}

Since many animals have a rigid skeletal structure, we encourage that the length of active edges ($w_{m,n} > 0$ for point pairs $m$ and $n$) are consistent across samples. 
We do not assume that these lengths and connections are known, such as previous work~\cite{usman2022metapose}; rather, they are learned during training. 
We do this by maintaining a running average of the length of all active edges $l_{avg(m,n)}$, and minimizing the difference between the average length and each sample $l_{m,n}$:
\vspace{-0.05in}
\begin{align}
 \mathcal{L}_{length} = \sum_{m}\sum_{n} \mathbbm{1}_{w_{m,n} > 0} \left\Vert l_{avg(m,n)} -  l_{m,n} \right\Vert_2.
\end{align}

During training, we update $l_{avg(m,n)}$ using an exponential running average and $w_{m,n}$ indicating edge weights for every pair is learned. Both of these parameters are shared across all viewpoints and timestamps. Notably, the length constraint is only applied to active edges, since there are many point pairs without rigid connections (e.g. elbow to feet), while we want to enforce this constraint only for rigid connections (e.g. elbow to wrist). 

\vspace{-0.05in}
\subsubsection{Separation Loss}

To encourage unique keypoints to be discovered, we apply separation loss to our 3D keypoints, which has been previously studied in 2D~\cite{ZhangKptDisc18,sun2022self}. On a set of 3D keypoints $U_{it}$, where $i$ is the index of a keypoint and $t$ is the time, the separation loss is: 
\begin{align}
    \mathcal{L}_{s} = \sum_{i \neq j} \exp{\left( \frac{-(U_{it} - U_{jt})^2}{2\sigma_s^2} \right)},
\end{align}
where $\sigma_s$ is a hyperparameter that controls the strength of separation.

\subsubsection{Training Objective}\label{sec:final_objective}
Our full training objective is the sum of the multi-view spatiotemporal reconstruction loss $\mathcal{L}_{recon}$, learned length constraints $\mathcal{L}_{length}$, and separation loss $\mathcal{L}_s$:
\begin{align}
    \mathcal{L} = \mathcal{L}_{recon} + \mathbbm{1}_{epoch>e} (\omega_r \mathcal{L}_{length} + \omega_s\mathcal{L}_s).
    \label{eq:full_objective}
\end{align}
Our model is trained using curriculum learning~\cite{Bengio2009}. We only apply $\mathcal{L}_{length}$ and $\mathcal{L}_s$ when the keypoints are more consistent, after $e$ epochs of training using reconstruction loss. 

\section{Experiments}

We demonstrate BKinD-3D using real-world behavioral videos, using a human dataset and a recently released large-scale rat dataset (Section~\ref{sec:exp_setup}).
We evaluate our discovered keypoints using a standard linear regression protocol based on previous works for 2D keypoint discovery~\cite{JakabNeurips18,sun2022self} (also described in Section~\ref{sec:training_procedure}). 
Here, we present results on pose regression (Section~\ref{sec:results}) with ablation studies (Section~\ref{sec:ablation}), with additional results in supplementary materials.

\subsection{Experimental Setup}\label{sec:exp_setup}

\subsubsection{Datasets}

We demonstrate our method by evaluating it on two representative datasets: Human 3.6M and Rat7M. The datasets have different environments and focus on subjects of different sizes, with humans being about 1700mm tall and rats about 250mm long. 

\textbf{Human 3.6M}. We evaluate our method on Human3.6M to compare to recent works in self-supervised 3D from 2D~\cite{usman2022metapose}. 
Human 3.6M~\cite{ionescu2013human3} is a large-scale motion capture dataset with videos from 4 viewpoints. 
We follow the standard evaluation protocol~\cite{iskakov2019learnable,kocabas2019self} to use subjects 1, 5, 6, 7, and 8 for training and 9 and 11 for testing. 
Our test set matches the set specified in~\cite{usman2022metapose} using every 16th frame (8516 test frame sets).
Notably, unlike baselines such as~\cite{iskakov2019learnable}, our method does not require any pre-processing with 2D bounding box annotations but rather is directly applied to the full image frame. 

\textbf{Rat7M}. We also evaluate our method on Rat7M~\cite{dunn2021geometric}, a 3D pose dataset of rats moving in a behavioral arena. 
This dataset most closely matches the expected use case for our method, which is a dataset of non-human animal behavior in a static environment.
Rat7M consists videos from 6 viewpoints captured at 1328$\times$1048 resolution and 120Hz, along with ground truth annotations obtained from marker-based tracking. We train on subjects 1, 2, 3, 4, and test on subject 5, as in \cite{dunn2021geometric}. We train and evaluate on every 240th frame of each video (3083 train, 1934 test frame sets). 

\vspace{-0.1in}
\subsubsection{Model Comparisons}\label{sec:model_comparisons}

We compare our method with three main categories of baselines: supervised 3D pose estimation methods (ex:~\cite{iskakov2019learnable}), 3D pose estimation methods from 2D supervision (ex:~\cite{usman2022metapose}), and a 3D keypoint discovery method developed for control in simulation~\cite{chen2021unsupervised}. A more detailed comparison of methods in this space is in Table~\ref{tab:related_work}.
For baselines with model variations, we use evaluation results from the version that is the closest to our model (multi-view inference, and camera parameters during inference). We note that all previous methods require additional 3D or 2D supervision, or jointly training a reinforcement learning policy in simulation~\cite{chen2021unsupervised}, which we do not require for 3D keypoint discovery in real videos.
Another notable difference is that previous methods typically pre-process video frames using detected or ground truth 2D bounding boxes~\cite{iskakov2019learnable}, while our method does not require this pre-processing step.

Since 3D keypoint discovery has not been thoroughly explored, we additionally study methods in this area using multi-view 2D discovery and triangulation (Triang.+Reproj.), and multi-view 2D discovery with a depth map estimates (Depth Map), in addition to our volumetric approach (Section~\ref{sec:method}, BKinD-3D). 
For multi-view 2D discovery and triangulation, we use BKinD~\cite{sun2022self} to discover 2D keypoints in each view, and perform triangulation using camera parameters to obtain 3D keypoints. We then project the 3D keypoints for multi-view reconstruction. We add an additional loss on the reprojection error to learn keypoints consistent across multiple views.
For the depth map approach, in each camera view, we estimate 2D heatmaps corresponding to each keypoint alongside a view-specific depthmap estimate. The final 3D keypoints are then computed from a confidence-weighted average of each view's estimated 3D keypoint coordinates (from the per-view 2D heatmaps and depth estimates). 
More details on each method are in the supplementary materials.

\vspace{-0.1in}

\subsubsection{Training and Evaluation Procedure}\label{sec:training_procedure}

We train our volumetric approach using the full objective (Eq~\ref{eq:full_objective}). We scale images to $256\times256$ for training, with a frame gap of 0.4s for Human3.6M and 0.66s for Rat7M. We use a maximum volume size of 7500mm for Human3.6M and 1000mm for Rat7M. The results are computed for all 3D keypoint discovery methods with 15 keypoints unless otherwise specified. We train using videos from the train split with camera parameters provided by each dataset.

We evaluate our 3D keypoint discovery through keypoint regression based on similar methods from 2D, using a linear regressor without a bias term~\cite{sun2022self,JakabNeurips18,ZhangKptDisc18}. 
For this regression step, we extract our discovered 3D keypoints from a frozen network, and learn a linear regressor to map our discovered keypoints to the provided 3D keypoints in each of the training sets.
We then perform evaluation on regressed keypoints on the test set.

For metrics, we compute Mean Per Joint Position Error (MPJPE) in line with previous works in 3D pose estimation~\cite{iskakov2019learnable,iqbal2020weakly}, which is the L2 distance between the regressed and ground truth 3D poses, accounting for the mean shift between the regressed and ground truth points.
To compare to methods that require addition alignment before MPJPE computation (e.g.~\cite{usman2022metapose} which does not use camera parameters during inference), we also compute Procrustes aligned MPJPE (PMPJPE)~\cite{usman2022metapose,kocabas2019self,iqbal2020weakly}. 
PMPJPE applies the optimal rigid alignment to the predicted and ground truth 3D poses before metric computation.

\begin{table}
  \begin{center}
\scalebox{0.95}{
    \begin{tabular}{lccc}
        \toprule[0.2em]
        Method & Supervision & PMPJPE $\downarrow$ & MPJPE $\downarrow$ \\
        \toprule[0.2em]
        \multicolumn{4}{c}{\textbf{\textit{Supervised 3D}}} \\
        Anipose~\cite{karashchuk2021anipose} & 2D only& - & 33\\  
        Rhodin et al.~\cite{rhodin2018learning} & 3D/2D & 52 & 67\\        
        Isakov et al.~\cite{iskakov2019learnable} & 3D/2D & -  & 21\\
        \bottomrule[0.1em]
        \multicolumn{4}{c}{\textbf{\textit{Supervised 2D + self-supervised 3D}}} \\   
        CanonPose~\cite{wandt2021canonpose} & 2D & 53  & 74\\
        EpipolarPose~\cite{kocabas2019self} & 2D & 67  & 77\\
        Iqbal et al.~\cite{iqbal2020weakly} & 2D & 55 & 69\\
        MetaPose~\cite{usman2022metapose} & 2D & 32  & - \\ 
        \bottomrule[0.1em]
        \multicolumn{4}{c}{\textbf{\textit{3D Discovery + Regression}}} \\     Keypoint3D~\cite{chen2021unsupervised} & $\times$ & 168  & 368\\
        \textbf{Ours}: \\
        \quad Triang+reproj & $\times$ & 134  & 241\\    
        \quad Depth Map & $\times$ &122  & 161 \\
        \quad BKinD-3D & $\times$ & 105  & 125\\ 
        \bottomrule[0.1em]
    \end{tabular}}
  \caption{\textbf{Comparing performance with related work on Human3.6M}. We note that previous approaches typically require additional 2D or 3D supervision, whereas our model discovers 3D keypoints directly from multi-view video. The 3D keypoint discovery models are evaluated using a linear regression protocol (Section~\ref{sec:training_procedure}).}
  \label{tab:human36m}
  \end{center}\vspace{-0.8cm}
\end{table}

\subsection{Results}\label{sec:results}

We evaluate our discovered keypoints quantitatively using keypoint regression on Human3.6M (Table~\ref{tab:human36m}) and Rat7M (Table~\ref{tab:rat7m}). 
Over both datasets with diverse organisms, our approach generally outperforms all other fully self-supervised 3D keypoint discovery approaches.
Additionally, among all the approaches we developed for 3D keypoint discovery, BKinD-3D using the volumetric bottleneck performs the best overall. 
Results demonstrate that BKinD-3D is directly applicable to discover 3D keypoints on novel model organisms, potentially very different in appearance or size, without 2D or 3D supervision.  

Notably, on Humam3.6M, Keypoint3D~\cite{chen2021unsupervised}, developed for control of simulated videos, does not work well in our setting with real videos, and qualitative results demonstrate that this method was not able to discover keypoints that tracked the agent (supplementary materials).

\begin{figure*}
    \centering
    \includegraphics[width=\linewidth]{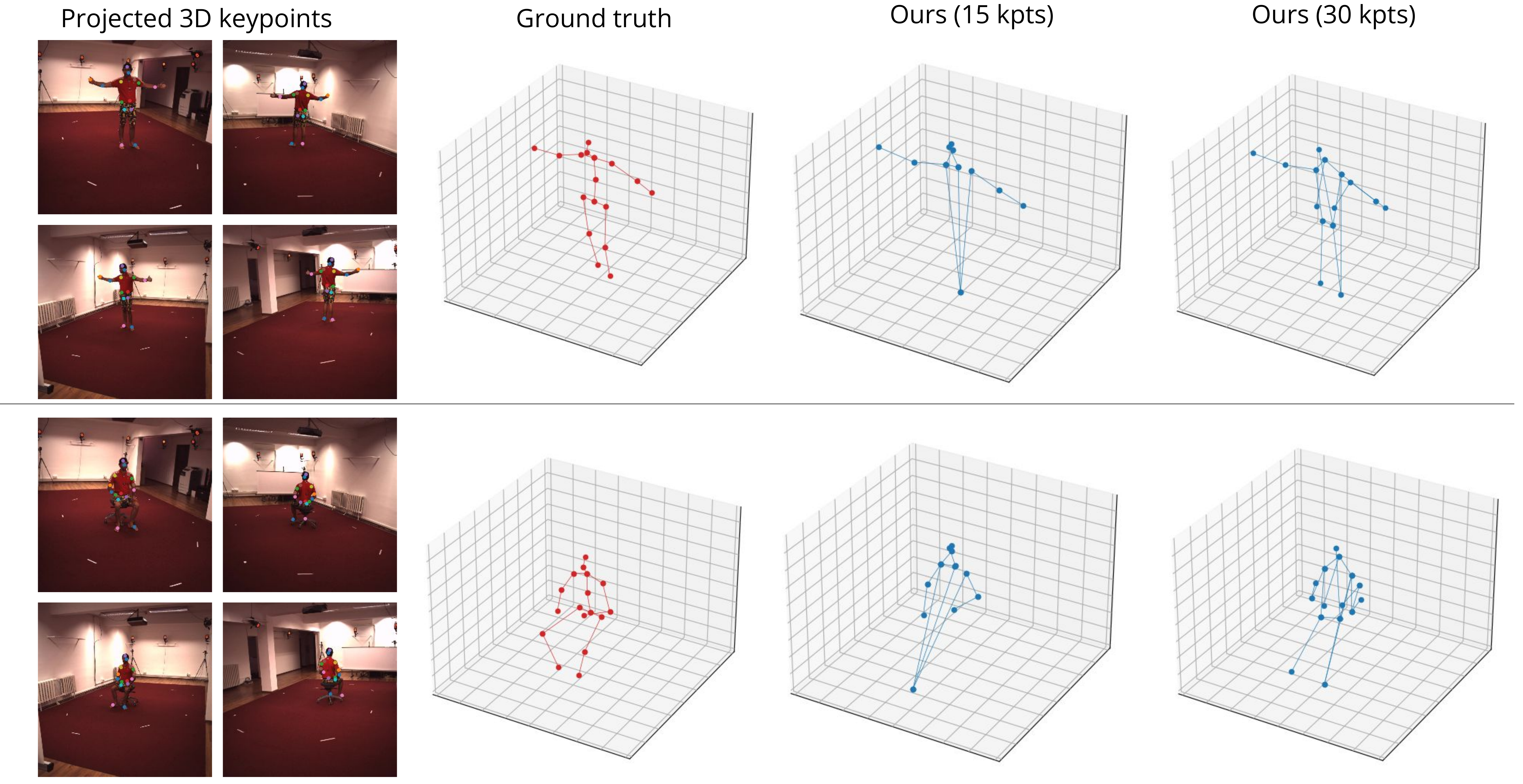}
    \caption{\textbf{Qualitative results for 3D keypoint discovery on Human3.6M}. Representative samples of 3D keypoints discovered from BKinD-3D without regression or alignment for 15 and 30 total discovered keypoints. We visualize all keypoints that are connected using the learned edge weights, and the projected 3D keypoints in the leftmost column are from the keypoint model with 30 discovered keypoints.
    }
    \vspace{-0.05in}    
    \label{fig:qualitative}
\end{figure*}

\begin{figure*}
    \centering
    \includegraphics[width=\linewidth]{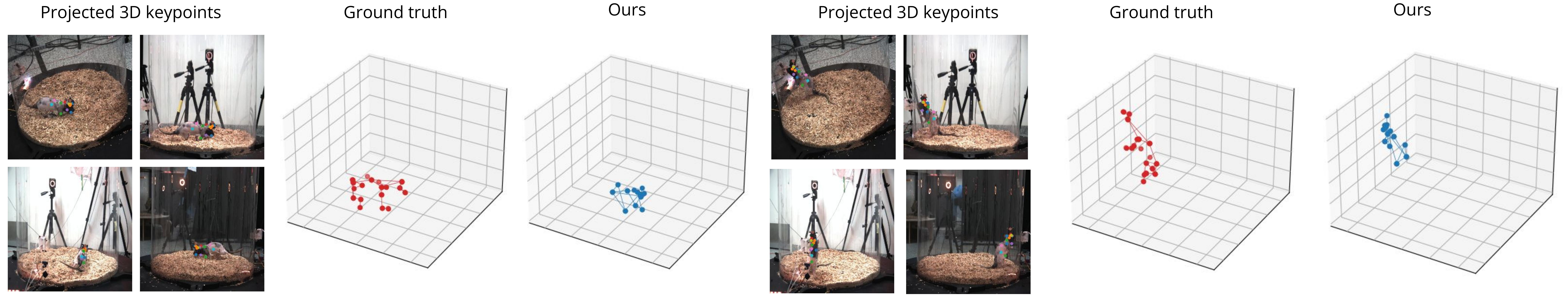}
    \caption{\textbf{Qualitative results for 3D keypoint discovery on Rat7M}. Representative samples of 3D keypoints discovered from BKinD-3D without regression or alignment. We visualize all connected keypoints using the learned edge weights and visualize the first 4 cameras (out of 6 cameras) in Rat7M for projected 3D keypoints.
    }
    \vspace{-0.1in}
    \label{fig:qualitative_rat}
\end{figure*}

\textbf{Qualitative results.} We find that the discovered points and skeletons are reasonable and look similar to the ground truth annotations for Human3.6M (Figure~\ref{fig:qualitative}) and Rat7M (Figure~\ref{fig:qualitative_rat}). Furthermore, we find that a volumetric model with 30 keypoints learns a more detailed human skeleton representation than a model with 15 keypoints. For example, the model with 30 keypoints is able to track both legs, while the 15 keypoint model only tracks 1 leg; however, both models miss the knees. Importantly, our model discovers the skeleton in global coordinates, and is able to track the agent as they move around the space. More examples are in supplementary materials.

While there exists a gap in terms of quantitative metrics between supervised methods and self-supervised 3D keypoint discovery, supervised methods require users to invest time and resources for annotations. In comparison, our method can be deployed out-of-the-box on new datasets and experiments with multi-view cameras.
Our approach has closed the gap substantially to supervised methods compared to previous work, without requiring time-consuming 2D or 3D annotations. 
Qualitative results demonstrate that our approach is able to discover structure across diverse model organisms, providing a method for accelerating the study of organism movements in 3D.

\textbf{Downstream Analysis.} To further evaluate our keypoint discovery method, we use BKinD-3D keypoints as input to a 1D convolutional neural network (previously used in~\cite{sun2021multi}) to predict action labels on Human3.6M. Notably, we found that our keypoints performs similarly to ground truth 3D points for action recognition, where Top 5 accuracy is 64.8\% (GT), 61.0\% (15 kpts), and 64.9\% (30 kpts) (supplementary material). 

\begin{table}
  \begin{center}
\scalebox{0.95}{
    \begin{tabular}{lccc}
        \toprule[0.2em]
        Method & Supervision & PMPJPE $\downarrow$ & MPJPE $\downarrow$ \\
        \toprule[0.2em]
        \multicolumn{4}{c}{\textbf{\textit{Supervised 3D}}} \\
        DANNCE~\cite{dunn2021geometric} & 3D & 11 & -\\        
        \multicolumn{4}{c}{\textbf{\textit{3D Discovery + Regression}}} \\   
        \textbf{Ours}: \\
        \quad Triang+reproj & $\times$ & 21 & 108\\
        \quad Depth Map & $\times$ & 27  & 56\\
        \quad BKinD-3D & $\times$ & 24  & 76 \\ 
        \bottomrule[0.1em]
    \end{tabular}}
  \caption{\textbf{Comparison with 3D keypoint discovery methods on Rat7M}. Results from the top three 3D keypoint discovery methods on Rat7M. The 3D keypoint discovery models are evaluated using a linear regression protocol (Section~\ref{sec:training_procedure}).}
  \label{tab:rat7m}
  \end{center}\vspace{-0.5cm}
\end{table}

\begin{table}
  \begin{center}
\scalebox{0.95}{
    \begin{tabular}{lccc}
        \toprule[0.2em]
        Method  & PMPJPE $\downarrow$ & MPJPE $\downarrow$\\
        \toprule[0.2em]
        BKinD-3D (8 kpts) & 120  & 149\\   
        BKinD-3D (15 kpts) & 105  & 125 \\ 
        BKinD-3D (30 kpts) & 109  & 130\\
        \hline
        BKinD-3D (point) & 110  & 137 \\         
        BKinD-3D (edge, without length) & 108 & 129 \\             
        BKinD-3D (edge, full objective) & 105 & 125\\             
        \bottomrule[0.1em]
    \end{tabular}}
  \caption{\textbf{Ablation results on Human3.6M}. We perform an ablation study of our volumetric bottleneck method comparing different numbers of keypoints as well as variations to the edge bottleneck with length constraints.
  }
  \label{tab:human36m_ablation}
  \end{center}\vspace{-0.5cm}
\end{table}

\vspace{0.05in}
\subsection{Ablation}\label{sec:ablation}
\vspace{0.05in}
We perform an ablation study of our model (Table~\ref{tab:human36m_ablation}), focused on BKinD-3D as it is the best performing approach on Human3.6M.
Results show that 15 keypoints performed the best quantitatively, but 30 keypoints is comparable and qualitatively provides a more informed skeleton (Figure~\ref{fig:qualitative}). 
We perform additional regression experiments using a 2-layer MLP regressor (supplementary material), and we found that the keypoints discovered by the 30 keypoints model (94 PMPJPE) perform better relative to 15 keypoints (98 PMPJPE). This suggests that the linear model may have been underfitting our 30 keypoints model.

We additionally find that adding edge information has a quantitative improvement on performance and provides more qualitative information on connectivity between joints (Figures~\ref{fig:qualitative}, ~\ref{fig:qualitative_rat}). In our 3D setting, we found that the point bottleneck (studied in previous works in 2D~\cite{sun2022self,JakabNeurips18}) did not work as well as the edge bottleneck (studied in previous works in 2D~\cite{he2022autolink}). By studying edge bottlenecks in 3D and expanding beyond 2D, our approach is able to enforce joint length constraints through the discovered 
edge connectivity.

\section{Discussion}

We present a method for 3D keypoint discovery directly from multi-view video, without any requirement for 2D or 3D supervision.
Our method discovers 3D keypoint locations as well as joint connectivity in behaving organisms using a volumetric heatmap with multi-view spatiotemporal difference reconstruction. 
Results show that our work has closed the gap significantly to supervised methods for studying 3D pose, and is applicable to different organisms.

Our approach focuses on behavioral videos with stationary cameras and background, with known camera parameters. 
The applicability of 3D keypoint discovery can be further improved with future work to jointly estimate camera parameters, camera movement, and pose from visual data. 
Additionally, the lack of publicly available multi-view datasets of animals could limit model development and evaluation. Open-sourcing more datasets in this area would encourage the development of pose estimation models with broader impacts beyond humans. 
Despite these challenges, 3D keypoint discovery has the potential to enable studying behavior of diverse organisms, without the need for expensive and time-consuming annotations. 
Our goal is to encourage more efforts in 3D keypoint discovery, to study the capabilities of vision models and to facilitate the study of behavior in new organisms and across diverse experimental setups.

\vspace{-0.05in}    
\section{Acknowledgements}

This work is generously supported by the Amazon AI4Science Fellowship (to JJS), NIH NINDS (R01NS102333 to JCT), the Air Force Office of Scientific Research (AFOSR FA9550-19-1-0386 to BWB), and NSF (1918865 to YY).

{\small
\bibliographystyle{ieee_fullname}
\bibliography{egbib}
}

\clearpage

\onecolumn

\appendix

\onecolumn

\section*{Supplementary Material}

We present additional discussions (Section~\ref{sec:additional_dis}), additional experimental results (Section~\ref{sec:addtional_results}), method description for the approaches we studied for 3D keypoint discovery in addition to the volumetric method (Section~\ref{sec:addtional_approaches}), additional implementation details (Section~\ref{sec:additional_implementation}), and qualitative results (Section~\ref{sec:qualitative_results}).
Our code is available at \url{https://github.com/neuroethology/BKinD-3D}.

\section{Additional Discussion}~\label{sec:additional_dis}
\textbf{Limitations and Future Directions}. 
Currently, our approach uses multi-view videos with camera parameters for training and focuses on behavioral videos with stationary cameras and backgrounds. 
Future directions to jointly estimate camera parameters, camera movement, and pose from visual data can improve the applicability of 3D keypoint discovery. 
We were also limited by the small amount of publicly available multi-view datasets of non-human animals. 
More open datasets in this space would encourage the development of pose estimation models with broader impacts beyond humans.
Finally, during our model training, once an edge between points becomes non-activated, then it is not displayed in the edge heatmap. To activate additional keypoints, one approach could be to perform random dropout on learned features of keypoints without activated edges and reset learned edge weights during training to activate additional keypoints.
While challenges exist, we highlight the potential for 3D keypoint discovery in studying the 3D movement of diverse organisms without supervision.

\textbf{Broader Impacts}. 3D keypoint discovery has the potential to accelerate the study of agent movements and behavior in 3D~\cite{marshall2022leaving}, since these methods does not require time-consuming manual annotations for training.
Additionally, behavioral scientists have long used summarizations of an animal’s skeleton in order to analyze behavior, as tracking the full skeleton is often impractical or infeasible.
To classify behavior, they have relied on estimates of
center of mass trajectory, PCA features, or even raw image
frames~\cite{pereira2020quantifying}.
This advance enables scientists to study behavior in novel organisms and experimental setups, for which annotations and pre-trained models are not available.
However, risks are inherent in applications of behavior analysis, especially regarding human behavior, and thus important considerations must be taken to respect privacy and human rights.
In research, responsible use of these models involves being informed and following policies, which often includes obtaining internal review board (IRB) approval, as well as obtaining written informed consent from human participants in studies.
Overall, we hope to inspire more efforts in self-supervised 3D keypoint discovery in order to understand the capabilities and limitations of vision models as well as enable new applications, such as studying natural behaviors of organisms from diverse taxa in biology. 

\section{Additional Experimental Results}~\label{sec:addtional_results}
We perform additional experiments of BKinD-3D on Human3.6M and Rat7M using our keypoint discovery model, focusing on the volumetric approach. We evaluate our keypoints using the 3D keypoint regression procedure specific in the main paper, unless otherwise specified.

\subsection{Keypoint regression using MLP vs linear model}

We additionally evaluate the volumetric model trained on Human 3.6M using a 2-layer multilayer perceptron (MLP) for keypoint regression.
Our MLP network has 50 hidden units as our regressor. 
We train the regressor on the train subjects and evaluate on unseen subjects (Table~\ref{tab:mlp}), matching the procedure for linear regressor in the main paper.
Using the MLP regressor, we find that the keypoints discovered by the 30 keypoints model perform better relative to with 15 keypoints (in the evaluation of the main paper the regressor was a linear model). 
This suggests that the linear model may have been underfitting our 30 keypoints model.

\begin{table}[b]
  \begin{center}
\scalebox{1.0}{
    \begin{tabular}{lccc}
        \toprule[0.2em]
          &  PMPJPE (MLP)	$\downarrow$ & MPJPE (MLP) 	$\downarrow$ \\
        \toprule[0.2em]
        BKinD-3D (15 kpts) & 98  & 116 \\ 
        BKinD-3D (30 kpts) & 94  & 111 \\
        \bottomrule[0.1em]
    \end{tabular}}
  \caption{
  \textbf{MLP regressor results on Human3.6M}.}
  \label{tab:mlp}
  \end{center}
\end{table}

\subsection{Action recognition results}

We compare our discovered keypoints to ground truth 3D keypoints as input to a 1D Convlutional Network (previously used in \cite{sarafianos20163d}) to predict action labels on Human3.6M (Table~\ref{tab:action}).
For train and test split, we use the same subject split as the main paper (subject 1,5,6,7,8 for train, and subject 9, 11 for test), and extract keypoints for all frames to classify actions.
The action classification network is a 3-layer 1D convolution with a window size of 15 with hidden dimensions 128, 64, and 32.
We find that 30 keypoints perform better for action recognition than 15 keypoints, and in particular, performs comparably to ground truth 3D keypoints in this setting.

\begin{table}
  \begin{center}
\scalebox{1.0}{
    \begin{tabular}{lccc}
        \toprule[0.2em]
          &  Top 1 Accuracy	$\uparrow$ & Top 5 Accuracy	$\uparrow$ \\
        \toprule[0.2em]
        BKinD-3D (15 kpts) & 31.8  & 61.1 \\ 
        BKinD-3D (30 kpts) & 36.5 & 64.9 \\
        \midrule[0.05em]
        Ground truth 3D kpts & 43.5 & 64.8 \\
        \bottomrule[0.1em]
             \vspace{-0.3in}
    \end{tabular}}
  \caption{
  \textbf{Action recognition results on Human3.6M}.}
  \label{tab:action}
  \end{center}
\end{table}

\subsection{Varying number of cameras}

\textbf{During both training and inference}. On Human3.6M (Table~\ref{tab:supp_h36m}), we vary the number of cameras from 4 to 2, and compute the mean performance over all camera pairs. For the 4 camera experiment, we used all 4 cameras for training and inference, while for the 2 camera experiment, we used the same selections of 2 cameras for training and inference. 
The mean performance with 2 cameras is slightly lower than using all cameras.
Notably, on the best performing camera pair, we observe that the performance is similar to using all 4 cameras. This result is promising for 3D keypoint discovery in settings that might limit the number of cameras, such as due to cost of additional cameras, maintenance effort, or difficulty of hardware setups. 

\textbf{During inference only}. We vary the number of camera during inference only, training on all 4 cameras and using pairs of 2 cameras for inference (Table~\ref{tab:supp_h36m}). We compute the mean, max, and min performance over all camera pairs. We note that this 2 camera inference setup performed slightly better compared to training on camera pairs and performing inference on the same 2 cameras. These results suggests that in experimental setups where cameras might fail or removed during recording, it may still benefit the model to be trained on all views, but then inference can be performed without re-training on the remainig views.

\subsection{Error distribution across joints}
\begin{figure*}
    \centering
    \includegraphics[width=0.48\linewidth]{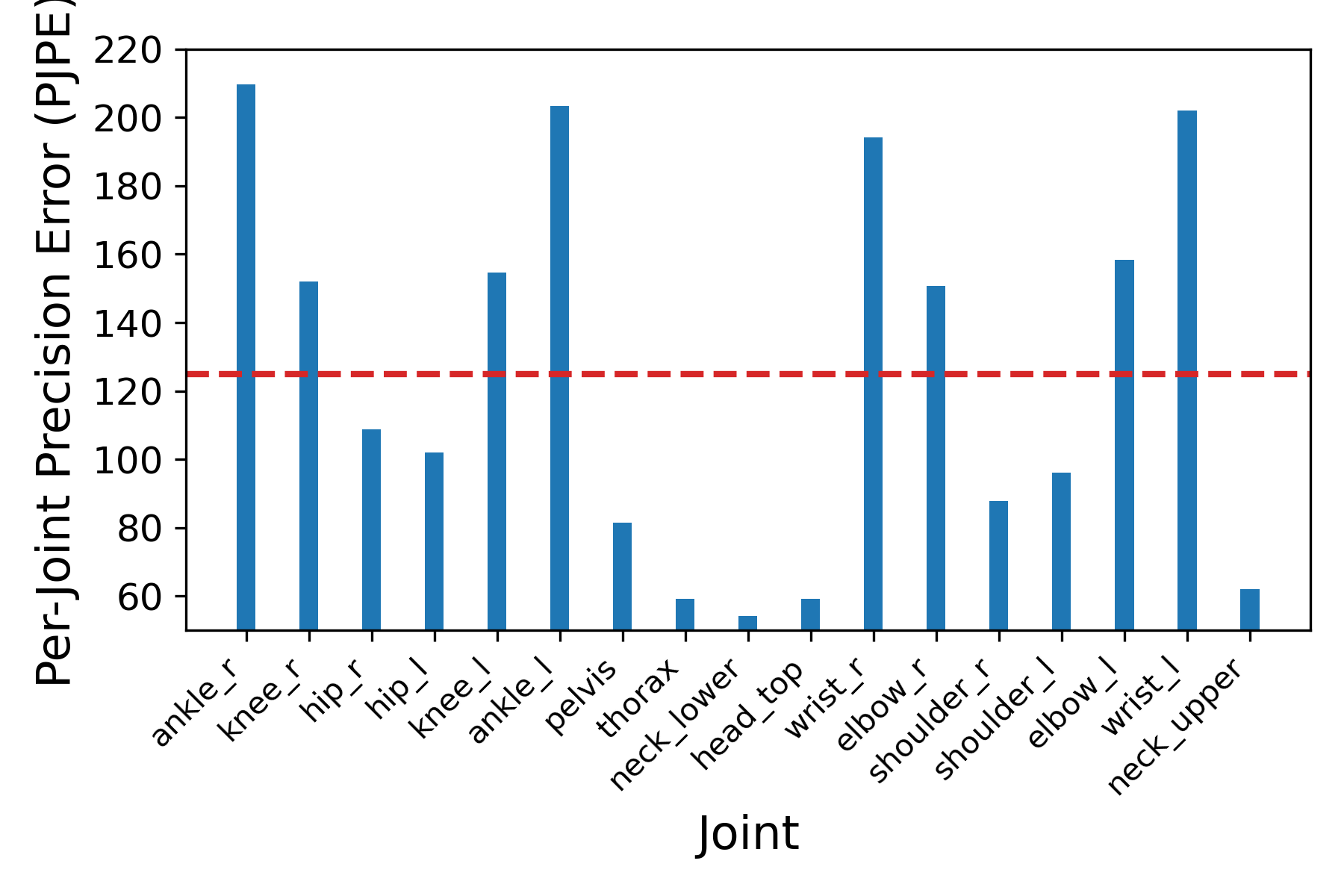}
    \includegraphics[width=0.48\linewidth]{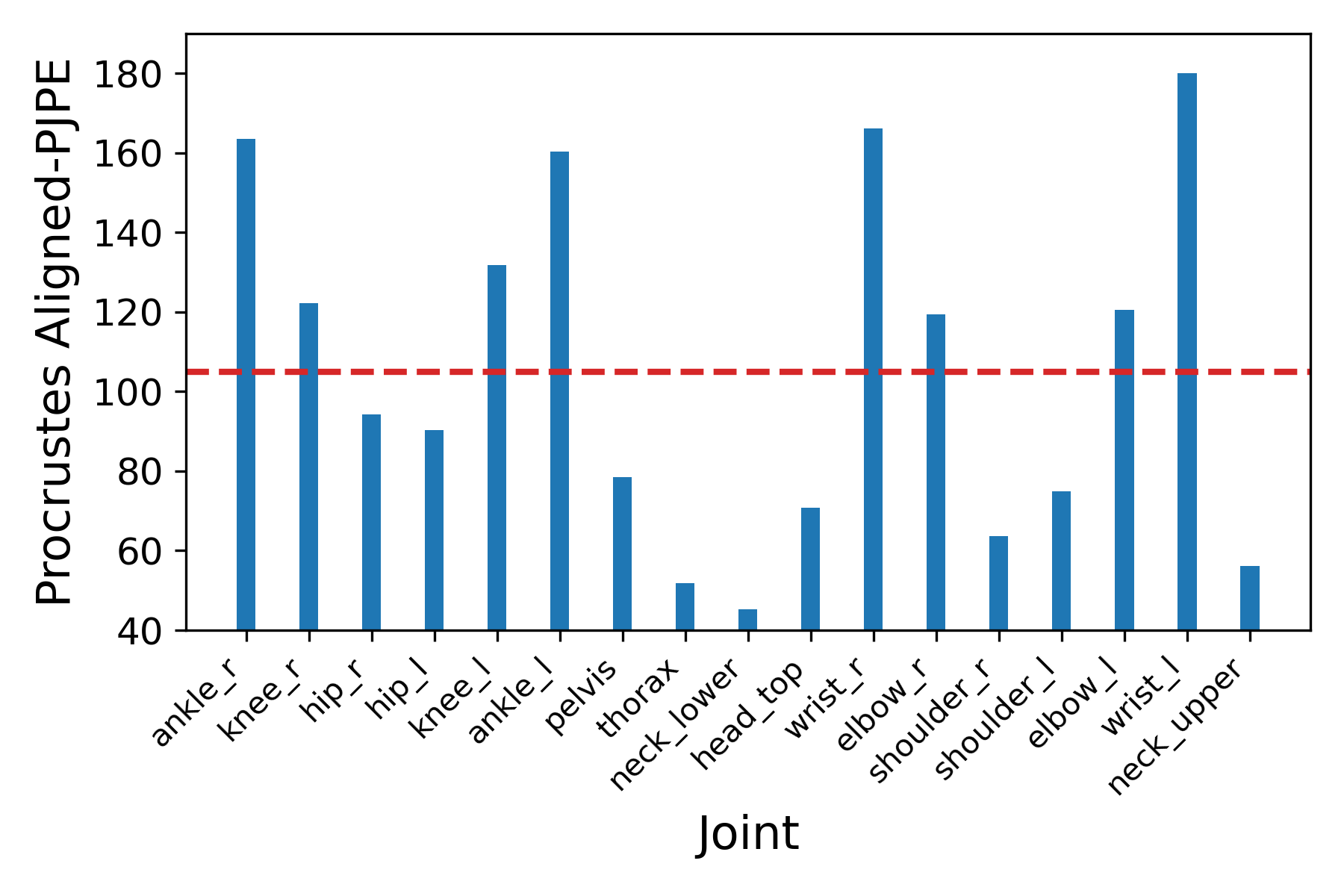}    
    \caption{\textbf{Per joint errors on Human3.6M}. Errors of each joint in mm using BKinD-3D, corresponding to the skeleton definition from the Human3.6M dataset. The dotted red line corresponds to the mean across joints (the MPJPE and PMPJPE respectively).
    }
    \label{fig:joints}
\end{figure*}

We visualize the error distribution across joint types from our 3D volumetric model (Figure~\ref{fig:joints}). We observe that generally joints on the limbs (e.g. wrist, ankle) have higher errors than joints closer to the center of the body (e.g. thorax, neck), for both MPJPE and PMPJPE.
This could be due to the wider range of motion of these limbs compared to the center in Human3.6M. 
There is not a significant difference in error across joints on the left side or right side. 
Since we currently perform inference per frame, future work to incorporate temporal constraints, or extend our method to identify meshes without supervision, could reduce errors on the limbs.

\begin{table}[b]
  \begin{center}
\scalebox{1.0}{
    \begin{tabular}{lccc}
        \toprule[0.2em]
        Method  & PMPJPE $\downarrow$ & MPJPE $\downarrow$\\
        \toprule[0.2em]
        BKinD-3D (4 cams) & 105 & 125\\           
        \multicolumn{3}{c}{\textbf{\textit{Train and inference with 2 cams}}} \\    
        BKinD-3D (2 cams) mean & 117  & 155 \\    \quad\quad\quad\quad\quad (2 cams) best  & 108  & 133 \\
           \quad\quad\quad\quad\quad (2 cams) worst & 125 & 167 \\
        \bottomrule[0.1em]
        \multicolumn{3}{c}{\textbf{\textit{Train with 4 cams, inference with 2 cams}}} \\          
        BKinD-3D (2 cams) mean & 114  & 153 \\    \quad\quad\quad\quad\quad (2 cams) best  & 103  & 130 \\
           \quad\quad\quad\quad\quad (2 cams) worst & 121 & 160 \\        
        \bottomrule[0.1em]
    \end{tabular}}
  \caption{\textbf{Camera variations on Human3.6M}. 
  We vary the number of cameras used for training and inference, as well as during inference only (trained with 4 cameras). Since there are multiple choices of 2 camera configurations, we chose the mean, best, and worst performance metrics. }
  \label{tab:supp_h36m}
  \end{center}
\end{table}

\subsection{Training with different keypoint counts in Rat7M}

On Rat7M (Table~\ref{tab:supp_rat7m}), we compare model performance when discovering 15 keypoints and 30 keypoints. We observe a small improvement in PMPJPE and MPJPE with an increased number of discovered keypoints, and also observe that the discovered keypoints cover a greater portion of the rat body in qualitative results (Figure~\ref{fig:qualitative_supp_rat7m}). 
This is similar to our observations on varying keypoints on Human 3.6M (Figure 3 in the main paper). It is possible that further increasing the number of keypoints could lead to a better body representation. Future work that explores using more efficient models with a much higher number of learned keypoints could further improve performance.

\begin{table}[b]
  \begin{center}
\scalebox{1.0}{
    \begin{tabular}{lccc}
        \toprule[0.2em]
        Method  & PMPJPE $\downarrow$ & MPJPE $\downarrow$\\
        \toprule[0.2em]    
        BKinD-3D (15 kpt) & 24  & 76\\   
        BKinD-3D (30 kpt) & 23  & 70 \\   
        \bottomrule[0.1em]
    \end{tabular}}
  \caption{\textbf{Additional results on Rat7M}. 
  We vary the number of discovered keypoints. }
  \label{tab:supp_rat7m}
  \end{center}
\end{table}

\subsection{Other hyperparameter variations}

\textbf{Volumetric representation}. We evaluate the performance of our model when varying the size of the volumetric features (Table~\ref{tab:supp_vol}). We did not observe a significant difference in performance with a bigger volumetric representation. This volume feature corresponds to $C$, which is the number of channels of the volumetric representation before input to the volume-to-volume network $\rho$.

\begin{table}
  \begin{center}
\scalebox{1.0}{
    \begin{tabular}{lccc}
        \toprule[0.2em]
        Method  & PMPJPE $\downarrow$ & MPJPE $\downarrow$\\
        \toprule[0.2em]    
        BKinD-3D (32 volume features) & 105  & 125 \\
        BKinD-3D (64 volume features) & 107  & 125 \\         
        \bottomrule[0.1em]
    \end{tabular}}
  \caption{\textbf{Varying volumetric representation size on Human3.6M.} }
  \label{tab:supp_vol}
  \end{center}
\end{table}

\textbf{Varying $\sigma$ in separation loss}. The value of $\sigma$ in the separation loss (section 3.2.3) controls the separation of the learned keypoints.
For our experiments, we used a value of $\sigma=0.08$ based on previous approaches in 2D~\cite{sun2022self}.
To evaluate the effect of this hyperparameter, we trained the BKinD-3D model on Human 3.6M with 15 keypoints with varying values of $\sigma$, evaluating with both our standard linear model and the MLP model described above (Table~\ref{tab:sigma}).
We find that changing the value of $\sigma$ does not change the test error signficantly. Qualitatively, we did find that the recovered keypoints were more evenly spread out on the human body for lower values of $\sigma$ (corresponding to a greater effect of the separation loss).

\begin{table} 
  \begin{center}
\scalebox{1.0}{
    \begin{tabular}{lccccc}
        \toprule[0.2em]
          &  PMPJPE (linear) $\downarrow$ &  PMPJPE (MLP) $\downarrow$ & MPJPE (linear) $\downarrow$ & MPJPE (MLP) $\downarrow$ \\
        \toprule[0.2em]
        
        BKinD-3D ($\sigma=0.04$) & 106 & 98 & 128 & 118\\  
        BKinD-3D ($\sigma=0.08$) & 105 & 98 & 125 & 116 \\ 
        BKinD-3D ($\sigma=0.16$) & 106 & 100 & 128 & 122\\ 
        \bottomrule[0.1em]
    \end{tabular}}
  \caption{
  \textbf{Varying $\sigma$ in separation loss on Human 3.6M}.}
  \label{tab:sigma}
  \end{center}
\end{table}

\section{Additional 3D Keypoint Discovery Approaches}~\label{sec:addtional_approaches}

\subsection{Triangulation and reprojection}

One of the simplest approaches to extending current 2D keypoint discovery methods~\cite{sun2022self} to three dimensions is to triangulate the discovered 2D keypoints to obtain 3D keypoints, then reproject the points back to 2D. This model can be trained using the same loss (spatiotemporal difference reconstruction) using the discovered 2D keypoints and the projected 2D keypoint in each view. We implement this approach, along with an additional loss for minimizing reprojection error to encourage detecting consistent keypoints across views.

We use an encoder-decoder architecture, with a shared appearance encoder $\Phi$, geometry decoder $\Psi$, and reconstruction decoder $\psi$.  For each camera view $i$ and time $t$, a frame $I_{t}^{(i)}$ is processed to obtain a heatmap $H_{t}^{(i)} = \Psi(\Phi(I_{t}^{(i)}))$. We apply a spatial softmax to obtain 2D keypoints $y_{t}^{(i)}$ for each view. The 2D keypoints across all views are triangulated to produce 3D keypoints. The triangulation is done by applying singular value decomposition (SVD) to find a solution to the following problem:
\begin{align*}
    \text{argmin}_{\tilde{z}_{t}^{(i)}} ||y_{t}^{(i)} - P^{(i)} \tilde{U}_{t}||_2
\end{align*}
where $\tilde{U}_{t}$ represents the 3D keypoints in homogeneous coordinates  and $P^{(i)}$ the projection matrix for camera view $i$.
The 3D keypoints are projected back into 2D for each view forming $y*_{t}^{(i)}$. 

To train the network, we minimize a sum of three losses:
\begin{itemize}
    \item $\mathcal{L}_{recon}^{(i)}$: the multi-view reconstruction loss (described in Section 3.2.1 of the main paper) using the detected 2D keypoints $y_{t}^{(i)}$ and $y_{t+k}^{(i)}$ 
    \item $\mathcal{L}_{projrecon}^{(i)}$: the same multi-view reconstruction loss as above, but applied to the projected 2D keypoints $y*_{t}^{(i)}$ and $y*_{t+k}^{(i)}$ 
    \item $\mathcal{L}_{reproj}^{(i)} = ||y*_{t}^{(i)} - y_{t}^{(i)}||_2$: the reprojection error 
    \item $\mathcal{L}_{s}$: the separation loss (described in Section 3.2.3 of the main paper)
\end{itemize}

The final loss is 
$$\mathcal{L} =  \sum_i \mathcal{L}_{recon}^{(i)} +  \mathbbm{1}_{epoch>e}( \omega_s \mathcal{L}_s + \omega_{p} \sum_i \mathcal{L}_{projrecon}^{(i)} + \omega_{r} \sum_i \mathcal{L}_{reproj}^{(i)})$$

Our model is trained using curriculum learning~\cite{Bengio2009}. We only apply the losses based on projected 2D points after $e$ epochs, when the model learns some consistent keypoints with each view. We train our model for 5 epochs and apply the losses after $e=2$ epochs.

\subsection{Depth Approach}
\begin{figure*}
    \centering
    \includegraphics[width=0.9\linewidth]{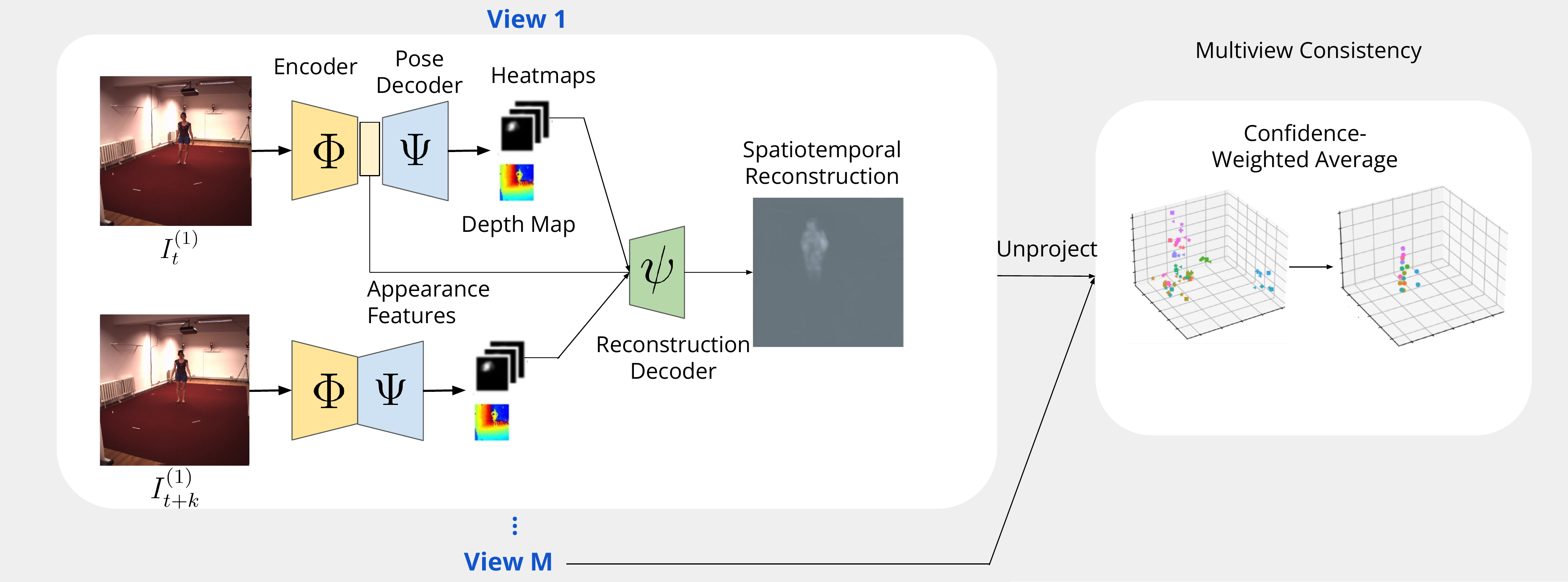}
    \caption{\textbf{3D keypoint discovery using depth maps}. The model is trained using multi-view spatiotemporal difference reconstruction to learn 2D heatmaps and depth representations at each view. Then the 3D information from each view is aggregated using a confidence-weighted average to produce the final 3D pose.
    }
    \label{fig:depth_method}
\end{figure*}

Based on the success in 2D unsupervised behavioral video keypoint discovery~\cite{sun2022self} and 3D keypoint discovery for robotic control~\cite{chen2021unsupervised}, we experiment with a framework that encodes appearance as well as 2D and depth representations (Figure~\ref{fig:depth_method}). Given multiple camera views with known extrinsic and instrinsic parameters, our framework learns 2D keypoints and depth maps to estimate 3D keypoints.

For each camera $i$, there is an appearance encoder $\Phi^{(i)}$, a pose decoder $\Psi^{(i)}$, and a depth decoder $D^{(i)}$. A frame $I_t^{(i)}$ and future frame $I_{t+k}^{(i)}$ are fed into the appearance encoder and subsequently the pose decoder. The pose decoder outputs $J$ heatmaps corresponding to the $J$ keypoints: $\Psi^{(i)}(\Phi^{(i)}(\cdot))$. A spatial softmax operation is applied to the output of the pose decoder, representing confidence or a probability distribution for the location of each keypoint. We interpret each of the heatmaps as a 2D Gaussian. The depth decoder outputs one depth map $D(\Phi(\cdot))$,  representing a dense prediction of distance from the camera plane for the scene. 
The appearance features $\Phi^{(i)}(I_t^{(i)})$ are fused with the 2D geometry features for both $I_t$ and $I_{t+k}$. These are fed into the reconstruction decoder $\psi$ to reconstruct the 2D spatiotemporal difference between $I_t$ and $I_{t+k}$. The spatiotemporal difference encourages the network to focus on meaningful regions of movement and be invariant to the background and other irrelevant features. This framework is repeated across camera views.

The reconstruction objective uses spatiotemporal difference reconstruction similar to our volumetric approach. To make the model more robust to rotation, we rotate the geometry bottleneck $h_g$ for image $I$ to create pseudo labels $h_g^{R^{\circ}}$ for the rotated input images $I^{R^\circ}$. where $R = {90^{\circ}, 180^{\circ}, 270^{\circ}}$. We apply mean squared error between the predicted geometry bottlenecks $\hat{h}_g$ and the rotated images and the generated pseudo labels $h_g$:
\begin{equation}
    \mathcal{L}_{rot} = \text{MSE} (h_g^{R^{\circ}}, \hat{h}_g(I^{R^\circ}) )
\end{equation}
The rotational loss can lead to a degenerate solution, with the keypoints converging to the center of the image. As such, we employ a separation loss as was done in our volumetric method.

For camera $i$ and a 3D point $(x,y,z)$ in the world coordinate system, we can use the projection matrix $P^{(i)}$ to project the 3D point to camera $i$'s normalized coordinate system $(u,v,d)$. Let the $\Omega^{(i)}$ operator denote the transformation to the camera plane and $\Omega^{*(i)}$ denote the inverse transformation. These transformations are differentiable and can be expressed analytically~\cite{chen2021unsupervised}.\\

After outputting the 2D keypoint heatmap $\Psi(\Phi(\cdot))$ and the depth map $D(\Phi(\cdot))$ for an input frame, we integrate over the probability distributions on the $\mathbb{R}^{S\times S}$ heatmaps and the depth maps to get the expected value for each coordinate $j$ and camera $i$: 
\begin{equation}
    \mathbb{E}[u_j^{(i)}] = \frac{1}{S}\sum_{u,v} u\cdot H_j^{(i)}(u,v)
\end{equation}
\begin{equation}
    \mathbb{E}[v_j^{(i)}] = \frac{1}{S}\sum_{u,v} v\cdot H_j^{(i)}(u,v)
\end{equation}
\begin{equation}
    \mathbb{E}[d_j^{(i)}] = \sum^S_{u=1}\sum^S_{v=1} D_j^{(i)}(u,v)\cdot H_j^{(i)}(u,v)
\end{equation}
The keypoints are unprojected into the world coordinate system:  $\Omega^-1_n(u,v,d)$. To penalize disagreement between predictions from different views, we use a multi-view consistency loss via mean-squared error.

\begin{table}
    \begin{center}
        \begin{tabular}{lccc}
        \toprule
            Type & Input dimension & Output dimension & Output size  \\
            \midrule
            Upsampling & - & - &  16x16 \\
            Conv\_block & 2048 + \# keypoints $\times$ 2 & 1024 & 16x16 \\
            Upsampling & - & - & 32x32 \\
            Conv\_block & 1024 + \# keypoints $\times$ 2 & 512 & 32x32  \\
            Upsampling & - & - & 64x64 \\
            Conv\_block & 512 + \# keypoints $\times$ 2 & 256 & 64x64  \\
            Upsampling & - & - & 128x128 \\
            Conv\_block & 256 + \# keypoints $\times$ 2 & 128 & 128x128  \\
            Upsampling & - & - & 256x256 \\
            Conv\_block & 128 + \# keypoints $\times$ 2 & 64 & 256x256 \\
            Convolution & 64 & 3 & 256x256 \\
            \bottomrule\\[-1em]
        \end{tabular}
    \caption{\textbf{Reconstruction decoder architecture.} ``Conv\_block" refers to combination of 3$\times$3 convolution, batch normalization, and ReLU activation. This architecture setup is also used for reconstruction decoding in~\cite{sun2022self,ryou2021weakly}.}
    \label{tab:decoder_details}
    \end{center}
\end{table}

\section{Additional Implementation Details}~\label{sec:additional_implementation}

\textbf{Architecture Details}. Our model architecture is based on ones studied before for 2D keypoint discovery~\cite{sun2022self,ryou2021weakly}.
Our encoder $\Phi$ is a ResNet-50~\cite{He2016DeepRL}, which outputs our appearance features.  Our pose decoder $\Psi$ uses GlobalNet~\cite{CPN17}, which outputs our 2D heatmaps. Our volume-to-volume network $\rho$ is based on V2V~\cite{moon2018v2v}. Finally, our reconstruction decoder $\psi$ is a a series of convolution blocks, where the architecture details are in Table~\ref{tab:decoder_details}. Our code is available at \url{https://github.com/neuroethology/BKinD-3D}.

\textbf{Hyperparameters}. The hyperparameters for the volumetric 3D keypoint discovery model  is in Table~\ref{tab:hyperparameter_1}. All keypoint discovery models are trained until convergence, with 5 epochs for Human3.6M and 8 epochs for Rat 7M. We use $\sigma = 0.08$ for the keypoint separation hyperparameter based on previous works~\cite{sun2022self}. We include additional details on each dataset:

\begin{table}[!t]
  \centering
  \small
  \scalebox{1.0}{
   \begin{tabular}{c | c | c| c | c| c  |c|c} 
   \hline
   Dataset & \# Keypoints & Batch size & Volume dimension & Volume size & Resolution & Frame Gap & Learning Rate  \\
   \hline
    Human3.6M & 15 & 1 & 7500 & 64 & 256 & 20 & 0.001\\
    Rat7M & 15 & 1 & 1000 & 64 & 256 & 80 & 0.001\\
   \hline
\end{tabular}
}
\caption{{\bf Hyperparameters for 3D Keypoint Discovery.} } \label{tab:hyperparameter_1}
\end{table}

\textbf{Human3.6M}.
The Human 3.6M dataset~\cite{ionescu2013human3} contains 3.6 million frames of 3D human poses with corresponding video captured from 4 different camera views, recorded from a set of different scenarios (discussion, sitting, eating, ...). Each scenario consists of videos from all 4 views with the same background, across a set of human participants. The person in the video is approximately 1700mm tall while the room is approximately 4000mm in dimension. The dataset is captured at 50Hz.
This dataset is licensed for academic use, and more details on the dataset and license are provided by the Human 3.6M authors within~\cite{ionescu2013human3}.

\textbf{Rat7M}. The Rat7M dataset~\cite{dunn2021geometric} consists of 3D pose and videos from a behavioral experiment with a set of rats, recorded across 6 views. 
This is currently one of the largest dataset with animal 3D poses.
The dataset consists of 5 rats, with videos from some of the rats across multiple days.
The rats are approximately 250mm long with the cage being around 1000mm in dimension.
The video is captured at 120Hz.
Some of the ground truth poses in Rat7M contains nans, and during processing, similar to ~\cite{dunn2021geometric}, we remove frames with nans from evaluation.
Our training procedure is not affected since we do not use any 3D poses during training.
This dataset is open-sourced for research.

\section{Qualitative Results}~\label{sec:qualitative_results}

We present additional qualitative results from BKinD-3D in Figures~\ref{fig:qualitative_supp_h36m} and \ref{fig:qualitative_supp_rat7m}.
For Human 3.6M (Figure~\ref{fig:qualitative_supp_h36m}), qualitative results demonstrate that the volumetric method discovers 3D keypoints and connections that qualitatively match the ground truth, even with self-occlusion or unusual poses, such as when the subject is laying or sitting down. 
The 30 keypoint model generally tracks the legs, shoulders, hips, arms, and head of the subject. The 15 keypoint model tracks the shoulders, arms, and head of the subject but fails to discover the legs and hips. This may be because we use spatiotemporal difference reconstruction, and there is more movements in these discovered parts. 
We observe that most discovered edges correspond to limbs, although there are extra discovered edges within the body. 
For example, the shoulders to feet connection in the 15 keypoint model. This edge likely allows the volumetric bottleneck to model the human shape with the limited keypoints available.
In addition to extra edges that may be discovered by our model, we may also miss parts, such as the knees of the subject, and occasionally the wrist keypoints (e.g. the left wrist for both 15 and 30 keypoint model in the last row).
Despite this, we note that the discovered skeleton is reasonable across a wide range of poses.

\begin{figure*}[b]
    \centering
    \includegraphics[width=\linewidth]{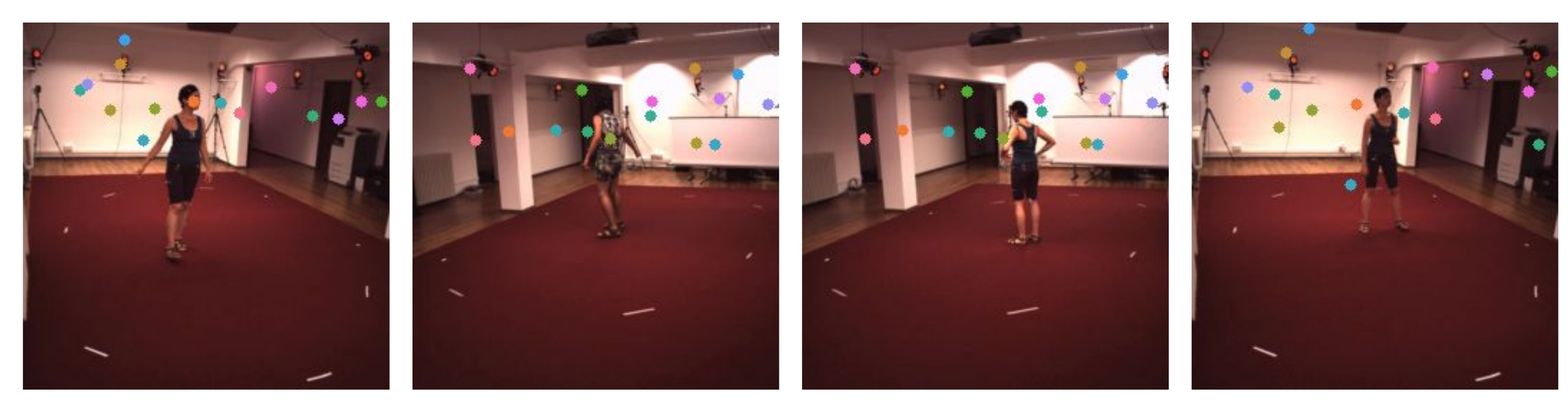}
    \caption{\textbf{Qualitative results for Keypoint3D on Human3.6M}. Representative samples of 3D keypoints discovered using Keypoint3D method~\cite{chen2021unsupervised} on real videos. 
    }
    \label{fig:keypoint3d}
\end{figure*}

\begin{figure*}
    \centering
    \includegraphics[width=0.8\linewidth]{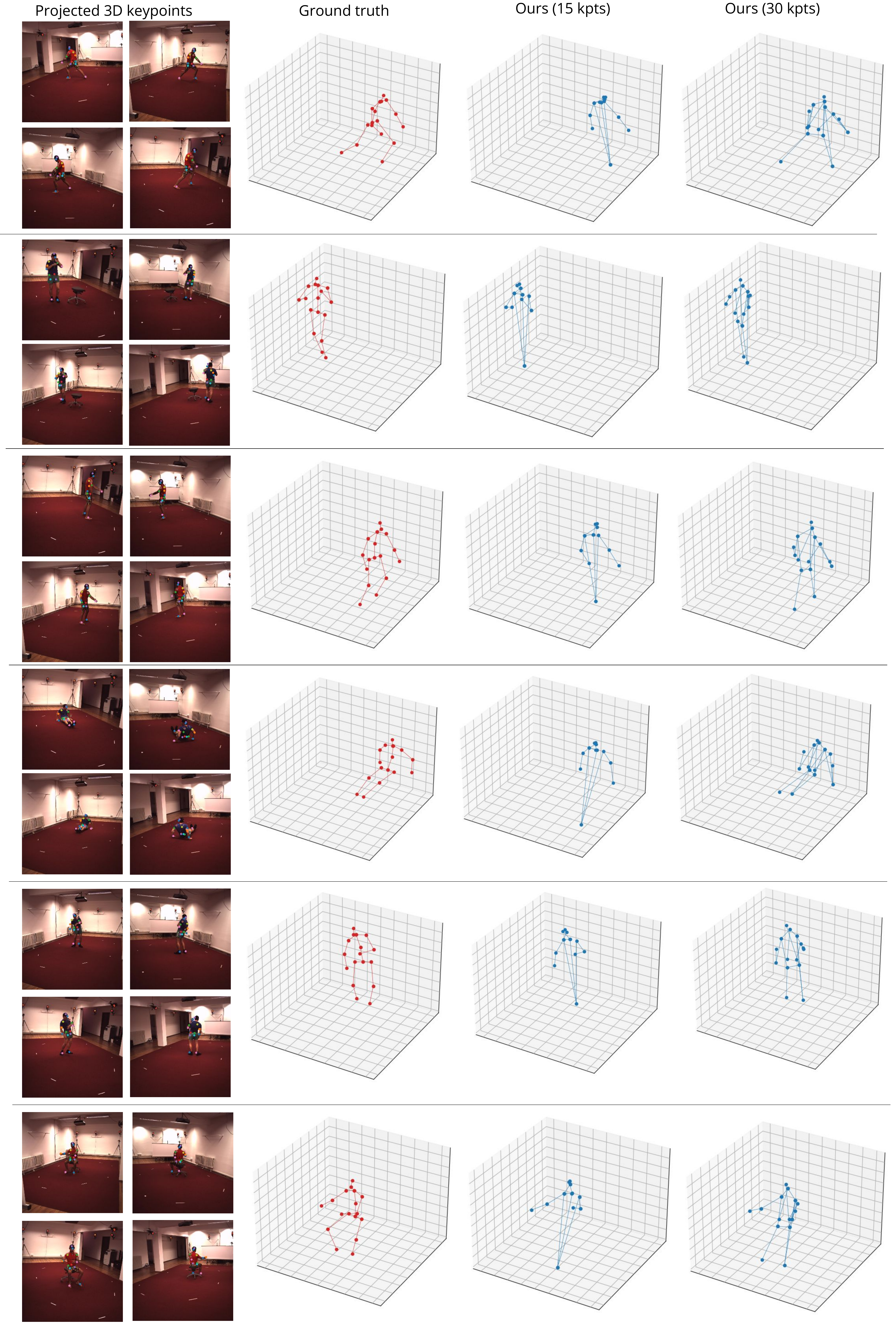}
    \caption{\textbf{Qualitative results for 3D keypoint discovery on Human3.6M}. Representative samples from BKinD-3D without regression or alignment for 15 and 30 total discovered keypoints. We visualize all keypoints that are connected using the learned edge weights, and the projected 3D keypoints in the leftmost column are from the keypoint model with 30 discovered keypoints.
    }
    \label{fig:qualitative_supp_h36m}
\end{figure*}

\begin{figure*}
    \centering
    \includegraphics[width=0.8\linewidth]{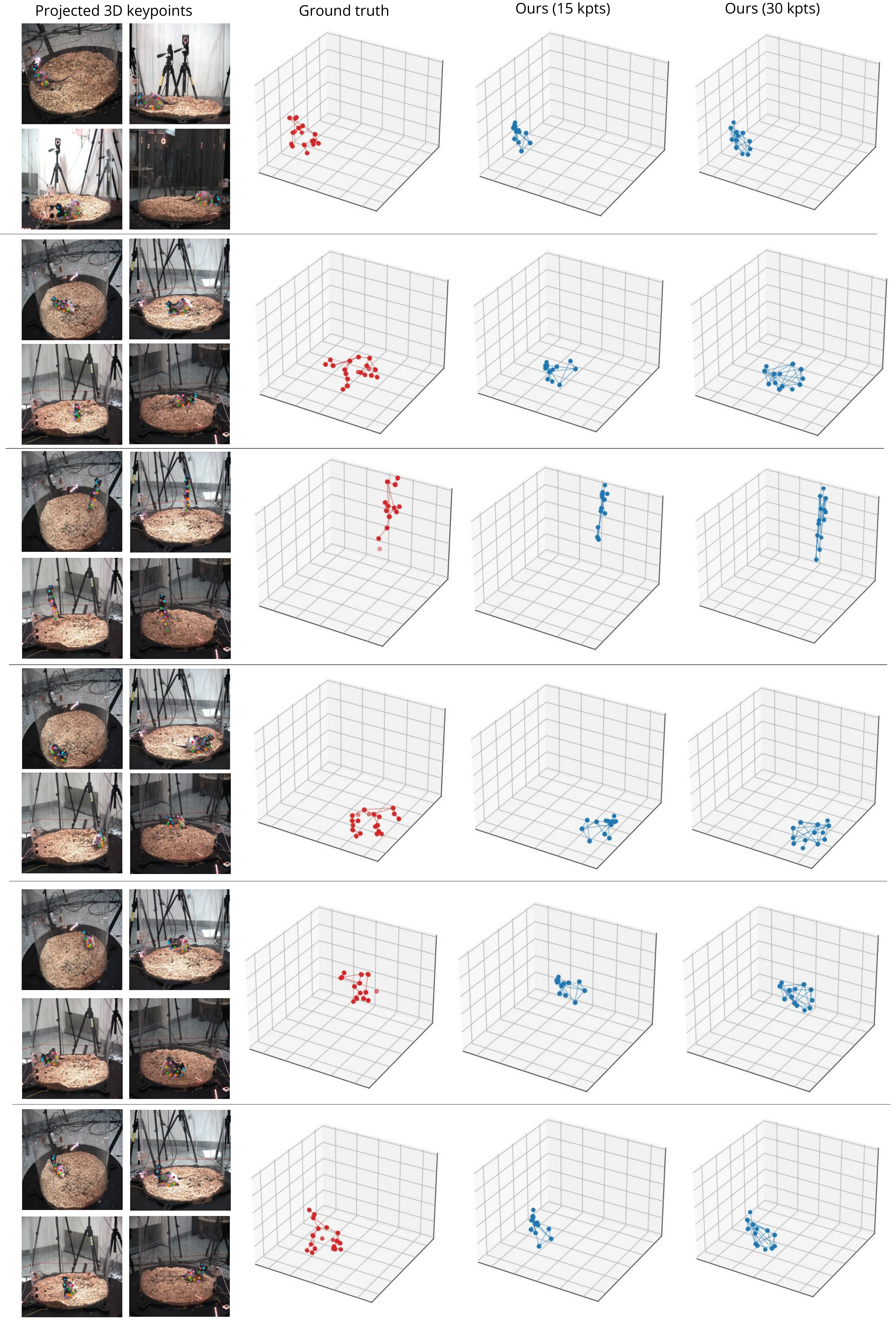}
    \caption{\textbf{Qualitative results for 3D keypoint discovery on Rat7M}. Representative samples of 3D keypoints discovered from BKinD-3D without regression or alignment for 15 and 30 total discovered keypoints. We visualize all connected keypoints using the learned edge weights and visualize the first 4 cameras (out of 6 cameras) in Rat7M for projected 3D keypoints from the 30 keypoint model.
    }
    \label{fig:qualitative_supp_rat7m}
\end{figure*}

In contrast to the volumetric bottleneck, the method Keypoint3D~\cite{chen2021unsupervised} does not work well on our real videos. In~\cite{chen2021unsupervised}, Keypoint3D jointly trains image reconstruction with a reinforcement learning (RL) policy loss in simulated environments. We find that training in real videos using only image reconstruction leads to poor performance: the discovered keypoints do not track any semantically meaningful parts (Figure~\ref{fig:keypoint3d}).

For Rat7M (Figure~\ref{fig:qualitative_supp_rat7m}), we also find that the volumetric bottleneck discovers interpretable keypoints that qualitatively match the ground truth. The head and front legs in particular are well tracked in both 15 and 30 keypoint models, across a variety of rat poses from having 4 feet on the ground to crouching to standing up. However, the back legs are only partially discovered in the 30 keypoint model. Furthermore, the discovered rat skeleton has much more edges compared to the ground truth. 
This highlights a limitation in our model, as the rat's skin and fat hide its underlying skeleton, making it difficult to discover the skeleton from video data alone. 
Future work could explore applying self-supervised learning constrained by body priors, such as animal X-rays, in order to discover a more precise skeleton.

Overall, qualitative results from the volumetric method demonstrates the potential of 3D keypoint discovery for discovering the pose and structure of different agents without supervision, across organisms that are significantly different in appearance and scale.

\end{document}